\documentclass[letterpaper, 10 pt, journal]{IEEEtran}%{ieeeconf}  % Comment this line out if you need a4paper

\IEEEoverridecommandlockouts                              % This command is only needed if 
\usepackage{amsmath} % assumes amsmath package installed
\usepackage{amssymb}  % assumes amsmath package installed
\usepackage{graphicx}
\usepackage{amsfonts}
\usepackage{bm}
\usepackage{nccmath}
\usepackage{hyperref}
\def\BibTeX{{\rm B\kern-.05em{\sc i\kern-.025em b}\kern-.08em
    T\kern-.1667em\lower.7ex\hbox{E}\kern-.125emX}}
    \hypersetup{
    colorlinks=true,
    linkcolor=blue,
    filecolor=magenta,      
    urlcolor=cyan,
}
\newcommand{\T}{{^\mathrm{T}}}

\newcommand{\R}{\mathbb{R}}

\newcommand{\bmat}[1]{\begin{bmatrix} #1 \end{bmatrix}}

\newcommand{\pmat}[1]{\begin{pmatrix} #1 \end{pmatrix}}

\newcommand{\p}{\partial}
\newcommand{\pdif}[2]{\frac{\p #1}{\p #2}}

\newcommand{\mpdif}[2]{\mfrac{\p #1}{\p #2}}

\mathchardef\mhyphen="2D
\newcommand{\inV}{{^{\mhyphen1}}}
\newcommand{\invT}{{^{\mhyphen\mathrm{T}}}}
\newcommand{\cin}{\!\in\!}

\newcommand{\dif}[2]{\frac{d #1}{d #2}}

\newcommand{\smashedunderbrace}[2]{\smash{\underbrace{#1}_{#2}}\vphantom{#1}}
\newcommand{\convhull}{\operatorname{convhull}}
\newcommand{\diag}{\operatorname{diag}}
\newcommand{\blkdiag}{\operatorname{blkdiag}}

\newcommand{\bM}{\mathbf}

\title{\LARGE \bf
The dynamic effect of mechanical losses of actuators on the equations of motion of legged robots
}

\author{Young{-}woo Sim and Joao Ramos% <-this % stops a space
\thanks{$^{1}$Authors are with the Department of Mechanical Science and Engineering at the University of Illinois at Urbana-Champaign, Urbana, IL 61801, USA. Corresponding author: {\tt\small sim17@illinois.edu}}%
}

\begin{document}

\maketitle
\thispagestyle{empty}
\pagestyle{empty}

%%%%%%%%%%%%%%%%%%%%%%%%%%%%%%%%%%%%%%%%%%%%%%%%%%%%%%%%%%%%%%%%%%%%%%%%%%%%%%%%
\begin{abstract}
% The internal friction of transmission does more harm than simply degrading torque and power delivery
% Internal power loss in the transmission limits the performance of a robotic system. This letter presents a variant of manipulator equation expressed in terms of mechanical efficiency of transmission. This approach is particularly beneficial to designers at the early stage of system design. A case study of leg design is presented as an application of this concept.

Industrial manipulators do not collapse under their own weight when powered off due to the friction in their joints. Although these mechanism are effective for stiff position control of pick-and-place, they are inappropriate for legged robots which must rapidly regulate compliant interactions with the environment. However, no metric exists to quantify the robot's perform degradation due to mechanical losses in the actuators. This letter provides a novel formulation which describes how the efficiency of individual actuators propagate to the equations of motion of the whole robot. We quantitatively demonstrate the intuitive fact that the apparent inertia of the robots increase in the presence of joint friction. We also reproduce the empirical result that robots which employ high gearing and low efficiency actuators can statically sustain more substantial external loads. We expect that the framework presented here will provide the foundations to design the next generation of legged robots which can effectively interact with the world.

% 1. big picture 
% The mechanical losses put legged robots into a trouble. Industrial robots that have bad efficiency and low-backdrivability shows bad behavior as a whole system for interaction with environment. We suggest that this phenomena, which originates from the frictional losses happening at the contacts internal to transmissions, has to be analyzed to achieve the tasks expected for legged robots. This letter provides a fundamental formulation to quantify the effect of power losses in the mechanical transmissions on the dynamics of the whole robotic system in the form of equation of motion. We quantitatively demonstrate that inertia of a robotic system increases with the presence of the friction. Also, using the mechanical efficiency of actuators, we show the robots payload changes whether the friction helps or resists.    
\end{abstract}

%%%%%%%%%%%%%%%%%%%%%%%%%%%%%%%%%%%%%%%%%%%%%%%%%%%%%%%%%%%%%%%%%%%%%%%%%%%%%%%%
\section{Introduction}

The mechanical losses in an actuator governs the system level dynamics of the robot. For instance, conventional industrial manipulators behave like a statue when they are powered off: they are \textit{non-backdrivable}. The characteristics which determine this behavior are the low efficiency and high friction in the gearbox. And although these mechanical transmission have been successfully utilized in industrial manipulators for stiff position control, they are not appropriate for compliant force control due to their high apparent inertia \cite{ImpactMitigation}. And hence, to enable legged robots to control their contact with the environment, one must analyze how the negative impact of low efficiency governs the robot performance. Towards this goal, this letter introduces a framework for studying how the energetic losses at joint-level propagate to the dynamic behavior of the robot at system-level.

\begin{figure}[t]
  \centering
  \includegraphics[width =1\linewidth]{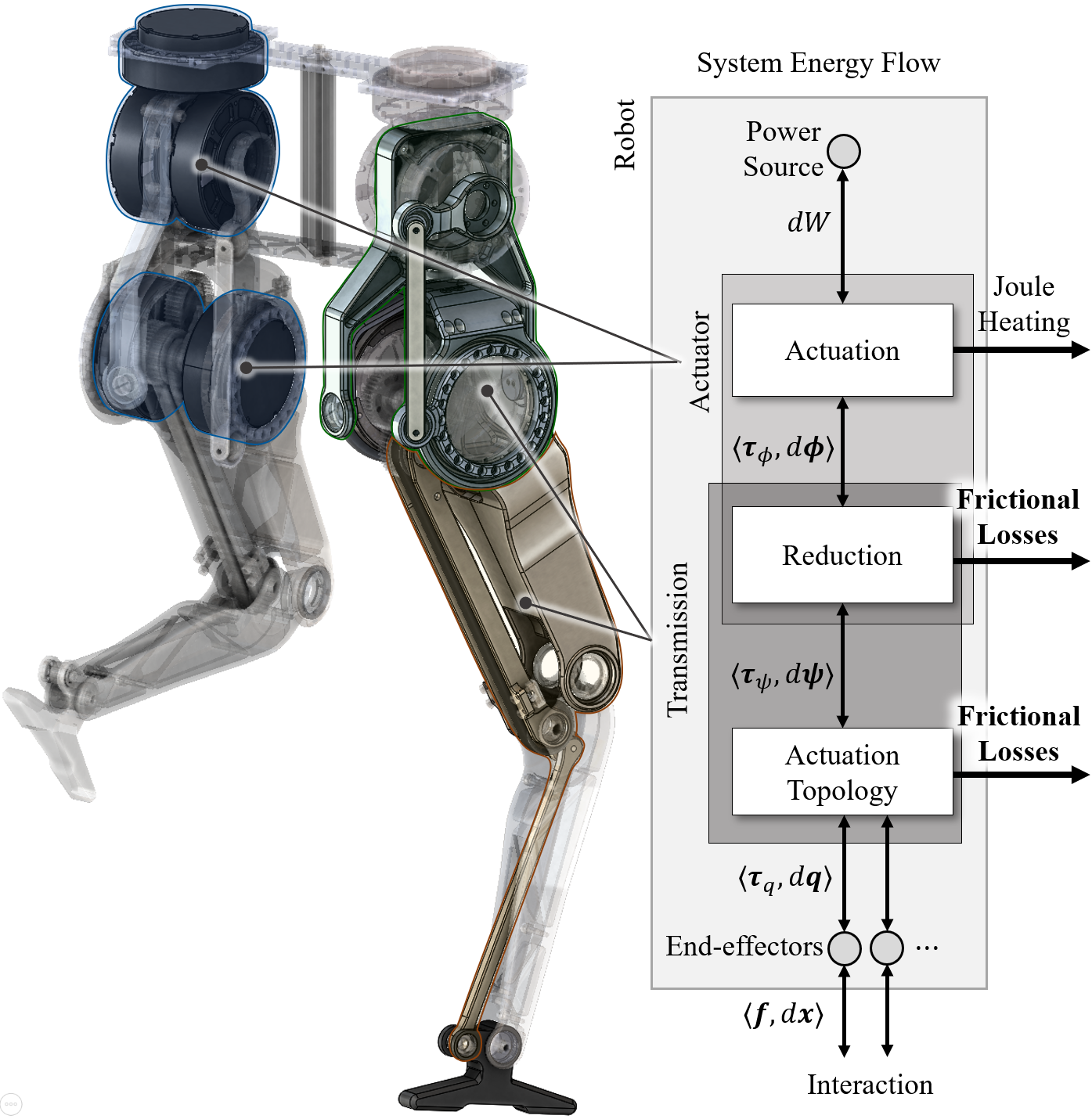}
  \caption{Left: The actuator placements and the actuation topology of a humanoid robot, TELLO. Right: Energy flow diagram of a general robot showing how actuators and transmissions dissipate energy. The actuators and transmissions of a robot function as mappings that convert the power flow in the system. These conversions are always accompanied by energy losses such as Joule heating or friction. }
  \label{Fig:TelloDiagram}
\end{figure}

No existing design metric can describe how the mechanical efficiency provided by manufacturers translates into a performance degradation of the robot. The absence of such metric prevents designers from utilizing the existing definition of mechanical efficiency to predict and compensate for the negative impact of mechanical losses \textit{a priori}. One must first build a robot and next experimentally estimate a model for the losses, which is specific to a particular mechanism embodiment. In addition, the lack of a metric which maps the actuator losses to the system-level dynamics hinders the selection of an optimal mechanical transmission for legged robots. For instance, no existing design guideline provides a clear choice between a compact and low efficiency strain wave gearbox and a bulkier, but higher efficiency, planetary gearbox. This unanswered question arose in the design of the lower-body of the humanoid robot \textit{TELLO}, shown in Fig. \ref{Fig:TelloDiagram}. and in \url{https://youtu.be/R0_2LmV3WQo}, and motivated this study. 
Related work in the literature analysed the fundamental behavior of mechanical transmission and their impact on actuator backdrivability. Giberti provides a guideline for the optimal choice of actuator-reducer pair, considering joint level efficiency \cite{ChoiceMotorReducer}, however, the analysis is limited to one degree-of-freedom (DoF). Similarly, Wang demonstrates that gearboxes present directional efficiency, which means that the mechanical losses are different if actuators operates within positive or negative work regimes \cite{Wang_DirectionalEfficiency}. Wensing investigates how joint-level apparent inertia decreases backdrivability, and thus, degrades the impact mitigation capability of the whole system \cite{ImpactMitigation}. Featherstone studies how the mass distribution of a robot's leg influences the propagation of impact from the ground to the torso \cite{SinghShockPropagation}. However, the impact of the mechanical efficiency of transmissions on the whole system still remains unstudied.

\begin{figure*}[t]
  \centering
  \includegraphics[width =1\linewidth]{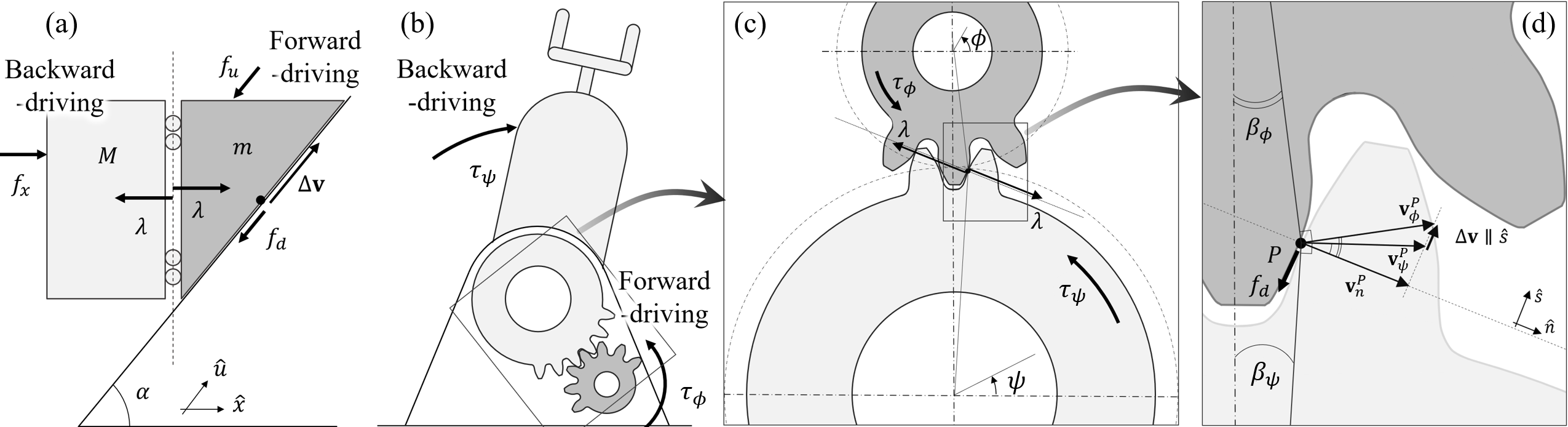}
  \caption{(a) A simple representative model analogous to a geared transmission. (b) The forward-driving and backward-driving of a robotic system. (c)(d) In depth views to trace the meshing force, $\lambda$, and the dissipative force, $f_d$. The frictional energy dissipation is associated with the slip, $\Delta \bm v$, at the contact contact point, $P$.}
  \label{Fig:SlopeAnalogy}
\end{figure*}

% contribution
The main contribution of this letter is to provide a fundamental formulation to quantify the effect of power losses in the mechanical transmissions on the dynamics of the whole robotic system. As a product of this formulation, we propose an augmented equation of motion that embeds the mechanical efficiency in the system's inertia and Coriolis forces matrices, and the vector of generalized forces. The system-level impact of the individual efficiencies is demonstrated by computing the force capability of the robot's end-effector and its generalized inertia ellipsoid \cite{AsadaInertiaEllipsoid}. Two core results are obtained from this study. First, the perceived inertia of the robot at the end-effector increases as the efficiency decreases. Second, the capability of the robot to generate or resist force depends on the direction of the power flow in the system.

% organization of the letter
This letter is organized as follows. First we study how transmission efficiency affects the dynamics of a one-DoF system in forward-driving (FWD) and backward-driving (BWD) scenarios of a representative example. Next, the generalized coordinates in robots and the mappings of the coordinates are introduced to define internal kinematic constraints. Moreover, the kinematic tree of rigid body system is highlighted to trace the energy loss inherent to the kinematic constraints. In section IV, the generalized dynamics of a rigid-body system with power losses in the kinematic structure is obtained. The proposed equation of motion enables the derivation of a {conventional} design criteria, the augmented generalized inertia ellipsoid, and the modified task-space force capability. Finally, a design study of a 2-DoF leg {qualitatively} validates the proposed formulation. 

% ==========================================================================
\section{A Simple Model of dissipative dynamics}
This section investigates how the power efficiency of a mechanical transmission is embedded into the equation of motion using a simple wedge-block model shown in Fig. \ref{Fig:SlopeAnalogy}(a). These dynamics extract the essential behavior of the complex sliding and pushing dynamics of the gear teeth meshing in the rotor-manipulator transmission model in Fig. \ref{Fig:SlopeAnalogy}(b). It is assumed that: (i) FWD occurs when pushing the wedge $m$ with a force $-f_u$ and, consequently, moving the block $M$. This is equivalent to commanding a motor torque $\tau_\theta$ to drive the link of a manipulator. And (ii) BWD occurs when power flows in the opposite direction by pushing the block with a force $f_x$ to drive the wedge. This is analogous to applying an external force to the manipulator's end-effector and generating a torque $\tau_\phi$ that drives the motors through the mechanical transmission.

The analogy between the wedge-block and gear meshing dynamics is described next. First, the geared coupling between the limb and the rotor in the manipulator joint is emulated by a holonomic constraint which relates the rotor angular displacement with the output angle of the gearbox
\begin{equation*}
   {g(\psi, \phi) = -\psi + \frac{1}{N}\phi  = 0.} 
\end{equation*}
Similarly, the wedge is constrained to move along the slope 
\begin{equation*}
    g(\bm q)= -x +{u}{\cos\alpha}=0.
\end{equation*}
where the generalized coordinate $\bm q = [x, \:u]\T$ of this example is given by the displacement of the block and the wedge, respectively. Consequently, the reduction ratio of the geared system, the ratio of input velocity to output velocity, is  $\frac{\dot{u}}{\dot{x}}=\tfrac{1}{\cos\alpha}$ in this example. Larger reduction ratios corresponds to larger slope angles as long as $\alpha < \tfrac{\pi}{2}$. 
Second, we study the dynamics of both FWD and BWD scenarios, where the meshing force, $\lambda$, and sliding friction, $f_d$, maintain their orientation and magnitude. However, the sign of the friction forces flips according to the movement of wedge, which contributes to the \textit{asymmetric} dynamic behavior in FWD and BWD. As a consequence, the example renders different mechanical efficiencies in the FWD and BWD scenarios, similarly to geared transmissions \cite{Wang_DirectionalEfficiency}. The power loss inherent to the gear meshing mechanics largely contributes to the asymmetry of the dynamics. Which means that the power loss, mechanical efficiency, apparent inertia, and input power distribution are different if the motors are driving the manipulator or if an external force is backdriving the actuators. However, we note that the mechanics of geared transmission are more complex than those in the wedge-block example. For instance, the relative velocity at the meshing surface of the gear is a periodic function of rotation angle, which makes the power loss a function the robot position and velocity. This phenomena is abstracted by estimating the average power loss over a rotation period. This concept of average power loss directly relates to the ordinary concept of constant mechanical efficiency of geared transmission \cite{YadaGearReview}. The rest of this section derives and discuss the dynamics of forward and back-driving cases.

\subsection{Constrained Dynamics and Meshing Force}
The core idea of this section is to combine the dissipative force $\bm f_d$ and the constraint force $\bM A \T \lambda$ in the Lagrangian formulation by lumping them as a net constraint force, namely the \textit{meshing force} $\bm r$. This is physically intuitive because the dissipative force $\bm f_d$ originates from the contact constraint $g(\bm q)$. In addition, the constraint Jacobian $\bM A =\pdif{g}{\bm q}=[-1,\: \cos\alpha]$ represents the mechanical advantage that distributes the constraint force. And thus, considering dry friction as the only dissipative force, the equation of motion is constructed using the Lagrangian formulation:
% The power loss induced by dissipative forces can be easily treated by combining constraint force, expressed by $\lambda$, and the dissipative force as a single term, namely, \textit{meshing force}, $\bm r$. In this example, dry friction is the only dissipative force. Backward-driving is defined as pushing the block to drive the wedge.
% The equation of motion is derived by the Lagrange multiplier method. The generalized coordinate is $\bm q=[x, u]\T$. 
% \vspace*{-3pt}
\begin{IEEEeqnarray*}{rcl}
\dif{}{t}\pdif{L}{\dot{\bm q}} - \pdif{L}{\bm q} - \bM A\T \lambda = \bm{f + f_d}, \\ 
\dif{}{t}\pdif{L}{\dot{\bm q}} - \pdif{L}{\bm q} - \underbrace{\left(\bM A\T \lambda + \bm{f}_d \right)}_{:=\bm r}= \bm{f }.
\end{IEEEeqnarray*}
%  \vspace*{-\baselineskip}
From which we obtain
% \vspace*{0.1\baselineskip}
% \vspace*{3pt}
\begin{IEEEeqnarray}{RL}
    \left[\begin{IEEEeqnarraybox*}[][c]{,c/c,}
                M & 0 \\ 0 & m\vphantom{\mu}
            \end{IEEEeqnarraybox*}\right]\!\!
    \left[\begin{IEEEeqnarraybox*}[][c]{,c,}
                \ddot{x} \\ \ddot{u}\vphantom{\mu}
            \end{IEEEeqnarraybox*}\right]
    \!-\! \underbrace{
    \left[\begin{IEEEeqnarraybox*}[][c]{,c,}
                -1 \\  \cos\alpha \pm \mu\sin\alpha \vphantom{\mu}
            \end{IEEEeqnarraybox*}\right]\!\lambda }_{(+):\bm{r}_f, \:\textrm{FWD }\:\:/\:\: (-):\bm{r}_b, \:\textrm{BWD}}
    &=\!
    \left[\begin{IEEEeqnarraybox*}[][c]{,c,}
                f_x \\ -f_u\vphantom{\mu}
            \end{IEEEeqnarraybox*}\right].\:\:   \label{eqn:exDyn}
\end{IEEEeqnarray} 
Equation \eqref{eqn:exDyn} demonstrates that the dry friction contributes to the asymmetric meshing forces $\bm{r}_f$ and $\bm{r}_b$. 

\subsection{Directionality and Efficiency}
\subsubsection{Frictionless Case} The classical Lagrange multiplier method assumes that the constraint force does not produce work. That is, constraint force and the generalized motion is always perpendicular to each other, and their inner product is always zero. Such motion is called \textit{tangent motion}, $d\bm q_t$, which belongs to tangent space defined by a constraint nullspace matrix, $\bM K$ s.t. $\bM{AK}=\bM 0$ \cite{TangentBundle}, \cite{LagrangianManifold}. Assuming the friction coefficient, $\mu$, is zero, the tangent motion and the work done by the meshing force are given by
\begin{gather*}
    d\bm q_t = \bM K dx, \\
    d W^{{\mu=0}}_{\bm r} = d\bm q_t\T \bm r = dx \bM K\T \bM A\T \lambda = 0,
\end{gather*}
where $\bM K = [1\: \sec\alpha]\T$. As a dual of the constraint Jacobian, the constraint nullspace matrix projects the general motion onto a subset of motion that is perpendicular to constraint force.

\subsubsection{Nonzero friction Case} The total work done by the meshing force including dry friction is always negative along the conventional tangent motion. 
\begin{IEEEeqnarray}{CCCCCC}
    dW_{\bm r} &=&{d\bm {q}_t}^{\mathrm{T}} \bm r &=& r_x dx + r_u du <0. \label{eqn:dZ}
\end{IEEEeqnarray}

To maintain mathematical consistency with the frictionless case (inner product equals zero), the inner product is formulated as the \textit{efficiency null}, $\delta Z$, which originates from the definition of mechanical efficiency $\eta$, {given by the ratio between output and input power}:
\begin{gather*}
    \delta Z = d\bm q_t\T \bM E \bm r= r_x dx + \eta r_u du = 0, \\
    \bM E =\bmat{1 & 0 \\  0 & \eta }\succ 0, \:\:
    \eta =\begin{cases}    \mfrac{1}{\eta_f} & \textrm{FWD}\\
                           \eta_b & \textrm{BWD}
           \end{cases},
\end{gather*}    
where $\eta_f$ and $\eta_b$ are the FWD and BWD efficiencies, and the efficiency matrix {$\bM E$} is positive definite and diagonal. The BWD and FWD efficiencies are derived from \eqref{eqn:dZ} by plugging in the meshing forces $\bm{r}_b$, and $\bm{r}_f$ from \eqref{eqn:exDyn} to obtain 
\begin{IEEEeqnarray*}{rCl}
    \eta_f &=& \frac{1}{1+\mu \tan \alpha}, \\[3pt]
    \eta_b &=& 1-\mu \tan\alpha.
\end{IEEEeqnarray*} 
The tangent motion for the nonzero friction case is:
\begin{equation}
d{\bm q}_t^{\mu} = \bM E d\bm q_t = \bM{EK} dx.    \label{eqn:MeshTangentMotion}
\end{equation}
The geometric intuition extracted from \eqref{eqn:MeshTangentMotion} is that the tangent space requires non-isometric stretch or compression, $\bM E$, according to the inclusion of dissipative force into the constraint forces.
% =====================================TABLE============================
\begin{table}
\renewcommand{\arraystretch}{1.3}
\caption{Constraint Space and Tangent Space of dissipative dynamics}
\label{table:ConstraintTangentSpace}
\centering
\begin{tabular}{c c c c}
\hline
 & \bfseries Constraint Space & \bfseries Tangent Space & \bfseries Inner Product\\
\hline
Scenario & {Force} & {Motion} & {Work}\\
{Frictionless} & $\bM A\T \lambda$& $\bM K dx$ & $0$\\
{Dissipative} & $\bM A\T\lambda + \bm{f}_d$ & $\bM E \bM K dx$ & $0$ \\
\hline
\end{tabular}
\end{table}
% =====================================TABLE============================
The modifications in the constraint and tangent spaces, and their duality are shown in the table. \ref{table:ConstraintTangentSpace}.

\subsection{Asymmetric Dynamics and Mechanical Impedance}
The efficiency null \eqref{eqn:dZ} is used for removing the Lagrange multiplier from the equation of motion. By multiplying $\bM K\T \bM E $ in both sides of the equality in \eqref{eqn:exDyn}, we obtain the dynamics of FWD and BWD cases:
\begin{IEEEeqnarray}{LL}
    \textrm{FWD:}& \quad \left( M + \eta_f \frac{m}{\cos^2 \alpha}\vphantom{\frac{1}{\eta_b}}\right)\ddot{x} = f_x-\eta_f\hat{f}_u
    \label{eqn:exDynFwdFinal},
\\[4pt]
        \textrm{BWD:}& \quad\left( M + \frac{1}{\eta_b} \frac{m}{\cos^2 \alpha}\right)\ddot{x} = f_x - \frac{1}{\eta_b}\hat{f}_u,
    \label{eqn:exDynBwdFinal}
\end{IEEEeqnarray}
where the force $f_u$ applied along the $\hat{u}$ coordinate frame is projected onto the $\hat{x}$ coordinate frame, or $\hat{f}_u = \tfrac{f_u}{\cos\alpha}$.

We utilize the Laplace transform to obtain the mechanical impedance $\bM X(s)$ in frequency domain $s$. We assuming that $\hat{f}_u$ is the only force applied in the FWD case and $f_x$ is the only force exerted in the BWD case.
% \begin{IEEEeqnarray}{crCl}
%     \textrm{FWD:}&\quad \bM{X}_f (s)=\frac{\bM x (s)}{\hat{\bM f}_u (s)} &=& \frac{1}{s^2}\left( \frac{1}{\eta_f}M + \frac{m}{\cos ^2 \alpha }\right)^{-1} \label{eqn:ForwardImpedance}\\[3pt]
%     \textrm{BWD:}&\quad \bM{X}_b (s)=\frac{\bM x (s)}{{\bM f}_x (s)}  &=& \frac{1}{s^2}\left( M + \frac{1}{\eta_b} \frac{m}{\cos ^2 \alpha }\right)^{-1} \label{eqn:BackwardImpedance}
% \end{IEEEeqnarray}

\begin{IEEEeqnarray}{crCl}
    \textrm{FWD:}&\quad \bM{X}_f (s)=\frac{\hat{\bM f}_u (s)}{\dot {\bM x} (s)} &=& \left( \frac{1}{\eta_f}M + \frac{m}{\cos ^2 \alpha }\right)s \label{eqn:ForwardImpedance}\\[3pt]
    \textrm{BWD:}&\quad \bM{X}_b (s)=\frac{{\bM f}_x (s)}{\dot{\bM x} (s)}  &=& \left( M + \frac{1}{\eta_b} \frac{m}{\cos ^2 \alpha }\right)s \label{eqn:BackwardImpedance}
\end{IEEEeqnarray}

\subsection{Discussion}
The results of this simplified setup addresses the fundamental asymmetry of dissipative dynamics. We show that the FWD and BWD dynamics are differently affected by friction $\mu$ and reduction ratio $\tfrac{1}{\cos\alpha}$. The unique properties of dissipative dynamics are summarized as follows:

\begin{itemize}
    \item \textit{Efficiency}: The backward efficiency is always smaller than forward efficiency. And both efficiencies are negatively affected by larger reduction and friction.
    \item \textit{Non-backdrivability}: There is a limiting case, {$\mu \tan\alpha=1$}, in which the system becomes non-backdrivable because $\eta_b=0$.
    \item \textit{Apparent inertia}: The efficiency increases the system's apparent inertia and the distribution of input forces. \textit{Both} the large gearing ratio and the low backward efficiency increases apparent inertia in the BWD case. 
    \item \textit{Impedance}: For both the FWD and BWD cases, the impedance of the system increases with the degradation of mechanical efficiency. 
\end{itemize}

These properties are also observed in the geared transmission as reported by researchers in \cite{BilateralGear, CycloidvsHarmonic, ImprovingBackdrivability}. Our results demonstrate that the overall BWD dynamics are {worse}. This occurs because, first, the apparent inertia of the wedge is inflated proportionally to the inverse of the backdriving efficiency. {Similarly, the negative impacts of low transmission efficiency are more likely to affect robots designed with lightweight limbs, in which the link inertia is dominated by the reflected inertia of motor rotor.} Second, he right hand side of \eqref{eqn:exDynBwdFinal} illustrates that any unmodeled friction in the transmission, such as stiction, is amplified by the reduction ratio \textit{and} the inverse of the backdriving efficiency to resist external forces, which degrades the backdrivability of the mechanism.

UPDATE: to mention impact mitigation, compare putting bad motors in different places. 

UPDATE: also talk about velocity dependence and damping -> not here. we can talk about this when coriolis term exists. 

%\section{Preliminary for Generalization}
\section{Generalization of the Dissipative Dynamics}
\label{sec_dissipativedyn}
{To generalize the dissipative dynamics to multi-DoF robotic systems, the generalized coordinates in robotic systems, and the mappings between the coordinates are first introduced. Next, these coordinate transformations are rearranged as a kinematic constraint that determines the mechanical connectivity of the robot. Next, we trace the frictional losses in the mechanical transmissions and discuss the topological propagation of power}

\subsection{Kinematic Topology and Constraints}

This section provides the mathematical definitions for two core concepts: the speed reduction and the actuation topology. Speed reduction is often realized by employing pairs of meshed gears or belt-driven pulleys with different diameters. The actuation topology describes the topology of the mechanism that distributes power from the motors to the joints. The actuation topology, or the kinematic structure \cite{SerialParallelComparison}, is used to classify robots as, for example, serial or parallel mechanisms. Readers can refer to \cite{SerialParallelComparison, GosselinSingularityAnalysis, MultiLoopKinematics} for the analysis of general nonlinear constraints. We define these two concepts as state transformations between coordinate angles. For that, three sequential states (angular positions) are introduced:
\begin{itemize}
    \item \textit{Rotor angle} $\bm \phi$: is the angular position of the motor rotor before a gearbox or a reduction mechanism;
    \item \textit{{Motor} angle} $ \bm \psi$: is the angular position of the output of the actuator after a  gearbox or a reduction mechanism; and
    \item \textit{Joint angle} $ \bm q $: is the angle between the structural links of the robotic system. 
\end{itemize}

And the mappings between the above mentioned states are defined as:
\begin{itemize}
    \item \textit{Reduction} $\bM G$: is the generalization of the speed reduction ratio, which maps the displacement of the rotor angle $\bm \phi$ to the motor angle $ \bm \psi$; and 
    \item \textit{actuation topology} $\bM D$: represents the {actuation topology} of the robot, mapping the motor angle $ \bm \psi$ to the joint angle $ \bm q $.
\end{itemize}

\begin{figure}[t]
  \centering
  \includegraphics[width =\linewidth]{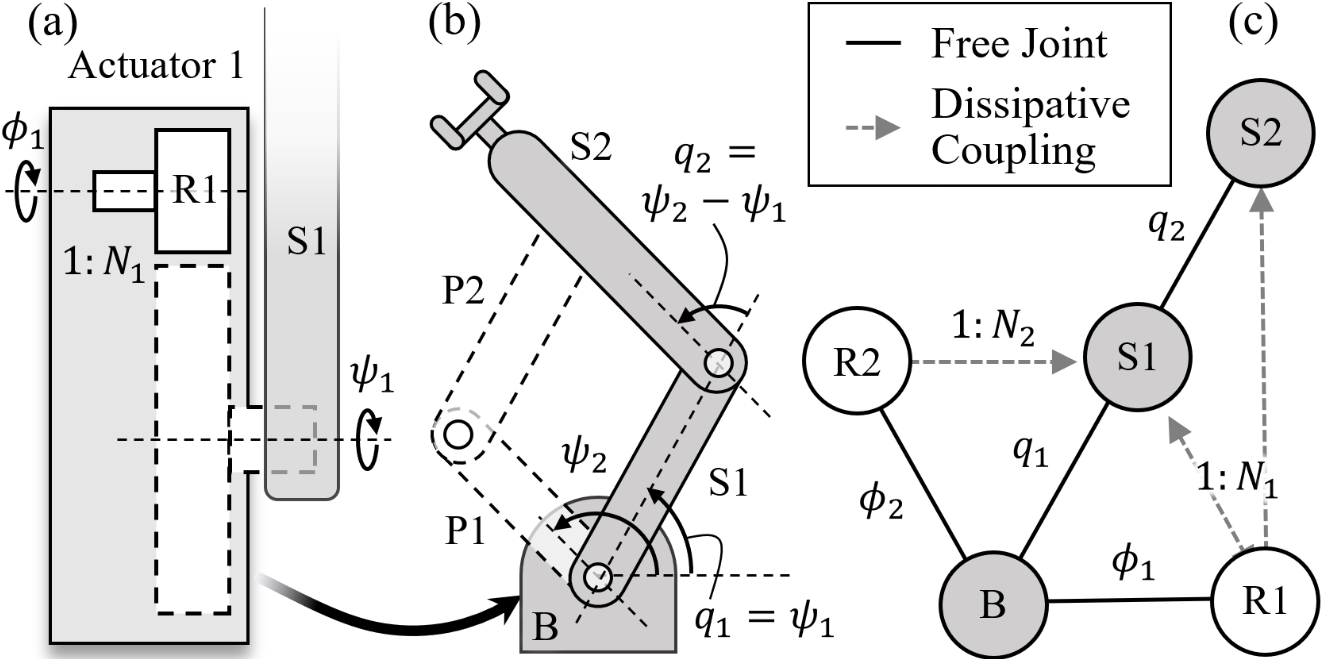}
  \caption{The coordinate angles in a 2-DoF robotic arm are visualized. The bodies with dashed boundaries are assumed to be massless. (a) A single stage gearbox in an actuator (b) A 2-DoF robotic arm using parallelogram (c) The kinematic tree outlines the mechanical topology and the kinematic constraints associated with power losses.}
  \label{Fig:RigidBody}
\end{figure}

Fig. \ref{Fig:RigidBody}(a) and (b) illustrates an {actuation topology} of a typical 2-DoF parallel robotic mechanism in which the speed reduction is represented by a single-stage gearbox. Similar mechanisms have been used in the legged robots ATRIAS \cite{ATRIAS_Design} and Minitaur \cite{Minitaur_DesignPrinciples}. The two motors are fixed to the base and drive the links $\textrm{S1}$ and $\textrm{P1}$ of the parallelogram mechanism. As a consequence, the input power propagates to the end-effector via links $\textrm{P2}$ and $\textrm{S2}$.
%A typical reduction mechanism and actuation topology, a single stage gearbox and a parallelogram, are provided in Fig. \ref{Fig:RigidBody}(a). and (b). Two motors are placed at the base $\textrm{B}$ and each of them drives a structural body $\textrm{S1}$ or body $\textrm{P1}$ that propagates force via $\textrm{P2}$ to $\textrm{S2}$. 
The gearbox reduces the rotation of the rotor $\textrm{R1}$ to motor angle $\psi_1=\tfrac{\phi_1}{N_1}$, where $N_1$ is the gearing ratio. The first joint is driven by the first actuator ($q_1=\psi_1$), while the second joint angle is controlled by both motors ($q_2=\psi_2-\psi_1$). Following our previous definitions, the reduction and actuation topology this manipulator are given by:
\begin{IEEEeqnarray*}{RCL} 
\bM G_{\rm par} = 
\left[\begin{IEEEeqnarraybox*}[][c]{,c/c,}
\tfrac{1}{N_1} & 0 \\
0 & \tfrac{1}{N_2}
\end{IEEEeqnarraybox*}\right],\quad \quad
\bM D_{\rm par} = 
\left[\begin{IEEEeqnarraybox*}[][c]{,c/c,}
1 & 0 \\
-1 & 1
\end{IEEEeqnarraybox*}\right].
\end{IEEEeqnarray*}
If the actuator topology represented a serial mechanism in which the motors are directly mounted on the joints, the matrix $D_{ser}$ would be equal to the $2\times2$ identify matrix, while the reduction matrix $G_{ser}$ would remain unchanged. These definitions are particularly useful to represent the internal holonomic constraints between rigid bodies. And thus, the general constraint null $\bm g$ defines the linearized relation between rotor angles and joint angles:
\begin{IEEEeqnarray}{rCl}
    \bm g(\bm q, \,\bm \phi) = \bm q - \bM{D G} \bm \phi. \label{eqn:genConstr}
\end{IEEEeqnarray}

Finally, we include the generalized coordinate of the end-effector task-space $\bm x$ and its Jacobian $\bM J$ to obtain the sequential transformations of generalized coordinates and their dual, the generalized forces: 
\begin{IEEEeqnarray}{CCCCCCC}
    d \bm \phi &\xrightarrow{\:\:\bM G\:\:} &d \bm \psi &\xrightarrow{\:\:\bM D\:\:}& d \bm q &\xrightarrow{\:\:\bM J\:\:}& d \bm x, \label{DualFlowMotion}\\
    d \bm{\tau_\phi}& \xleftarrow{\bM G\T}& d\bm{\tau_\psi} &\xleftarrow{\bM D\T}& d \bm{\tau_\phi}& \xleftarrow{\bM J\T}&
    d \bm{f_x}. \label{DualFlowForce}
\end{IEEEeqnarray}

To simplify the discussion, it is assumed that the matrices $\bM G$ and $\bM D$ are square, dimensionless, full-rank, and invertible. In addition, $\bM D$ is sparse and $\bM G$ is a diagonal matrix composed of the speed reduction ratios of the transmissions

\subsection{Rigid Body Systems and Kinematic Tree}
The work from Featherstone \cite{FeatherstoneBook} provides an efficient tool to visualize the connections between rigid bodies in a mechanism using kinematics trees. Following this concept, Fig. \ref{Fig:RigidBody}(c). shows the kinematic tree for the parallel mechanism described in the previous section.
This graph notation is useful for tracing the energetic losses associated with the kinematic constraints in the mechanism. We use this tool to define the rigid-body connections of a redundant robotic system using two types of connections: (i) free joints, which define the parent-child relations and (ii) dissipative couplings, which represent both kinematic constraint and the power losses associated with the constraint. In this sense, the dissipative couplings in Fig. \ref{Fig:RigidBody}(c). are represented by arrows to clearly illustrate that the power losses depend on the direction of power delivery.In order to avoid kinematic loops which would complicated the analysis, we assume that the power distribution mechanisms are massless. This assumption means that the components of the kinematic tree define the system's kinematics, but their inertia do not contribute to the system's equations of motion.

\section{The Dynamics of a Dissipative Rigid Body System}
The dissipative dynamics of a general multi-body robotic system is obtained similarly to the solution for the simple example using the definitions from section \ref{sec_dissipativedyn}. The overarching goal is to (1) create a kinematic tree with redundant rigid bodies (states), (2) define the sequential transformation using the linear holonomic constraints, (3) include the constraint forces, which include the dissipative forces, and (4) reduce the order of model by projecting its dynamics onto the tangent space using the efficiency null.  

The general model of a legged robot is comprised of a 6-DoF floating-base (the base or torso) and a serial chain of $m$-links (the limbs) [ref]. In addition, the limbs are driven by $m$ actuators, each with its own motor, and a mechanical transmission defined by the reduction matrix $\bM G\in \R^{m\times m}$. The actuator topology is defined by the matrix $\bM D\in \R^{m\times m}$. Therefore, the total number of DoF's of the redundant model is $6+2m$, while the reduced-order model is $(6+m)$-dimensional by including the kinematic constraints of the mechanism. The general robot model has $m$-pairs of coupled coordinates, $\bm \phi\in \R^{m}$ and $\bm q\in \R^{m}$, representing the rotor and joint angles, respectively. We define $\bm q_b\in \R^{6}$ as the 6-DoF coordinates for the free floating-base and write the vector of redundant states $\bm s$ and the reduced-order vector of coordinates $\bm y$ given by 
\begin{IEEEeqnarray*}{rCl}
    \bm s &=& [\bm q_b \T, \bm q \T, \bm \phi\T ]\T \in \R^{6+2m}, \\
    \bm y &=& [\bm q_b \T, \bm q \T ]\T \in \R^{6+m}. 
\end{IEEEeqnarray*}

The $m$ kinematic constraints between the rotors and the joints are defined by equation \eqref{eqn:genConstr}. The system constraint Jacobian $\bM A$ and its null-space $\bM K$ are 
\begin{IEEEeqnarray*}{rCl}
    \bM A &=& \mpdif{\bm{g}}{\bm s} = \bmat{\;\; \bm 0_{m\times6} & \bm1_{m}& -\bM{DG} \;\;} \in \R^{m\times(6+2m)},\\[4pt]
    \bM K &=& \bmat{\;\;\bm 1_6 & \bm 0 \\ \;\;\bm 0 & \bm 1_m \\ \;\; \bM 0 & (\bM{DG})^{-1}} \in \R^{(6+2m)\times (6+m)}.
\end{IEEEeqnarray*}
The projection or constraint null-space matrix,  $\bM K$, is selected from infinite solutions which satisfy $\bM{AK}=\bM 0$ because it preserves the reduced model's generalized coordinate 
\begin{IEEEeqnarray*}{rCl}
    d\bm y = \bM K d\bm s.
\end{IEEEeqnarray*}

The $m$ transmissions in the robot experience dissipative force $f_{d,j}$ and have individual efficiencies are $\eta_j$. We assumed that constraints between rotor and joint angles $g_j, j\in\{1,\cdots,m\}$ are independent and their efficiency nulls $\delta Z_j$ are equal to zero. Therefore, the meshing force, the total efficiency null and total efficiency matrices are
\begin{gather*}
    \bm r = \bM A\T \bm \lambda + \bm f_d, \\[4pt]
    \delta Z = \Sigma_{j=1}^{m} \delta Z_j = d\bm s\T \bM E \bm r =0,\\[4pt]
    \bM E = \bmat{\bm 1_6 & \bm 0& \bm0 \\ \bm 0 & \bm 1_m & \bm 0 \\ \bm 0 & \bm 0 & \bm \eta},
\end{gather*}
where $\bm f_d = [\bm{0}_{6+m}\T, f_{d,1}, \cdots , f_{d,m}]\T$ are the generalized dissipative forces, $\bm \lambda \in \R^{m}$ are the Lagrange multipliers and $\bm \eta = \diag(\eta_j)$  collects the individual transmission efficiencies $n_j$. 

The redundant system of equations of $s$ is projected onto the tangent space of $q$ by left multiplying $\bM{K\T E}$ on both sides of the equality of the equation of motion. The procedure eliminates the Lagrange multipliers from the expression and reduces the systems' dimension from $6+2m$ to $6+m$:  
\begin{gather}
    \bM{K\T E M K}\ddot{\bm y}  + \bM{K\T E}\bm c= \bM{K\T E}\bm{f}. \ \label{eqn:dissipativeDyn}
\end{gather}
The term  $\bm f \in \R^{6+2m}$ collects the joint torques $  {\bM{\bar J}}^{\mathrm{T}} {\bm f}_{\textrm{ext}} \in \R^{6+m}$ due to the contact forces $\bm f _{\textrm{ext}}\in \R^{6}$ at the end-effectors and the actuation torque $\bm \tau_{\phi} \in \R^{m}$. Given the contact Jacobian $\bM{\bar J} \in \R^{6\times(m+6)}$ and the Coriolis term $\bm c$. We assume there are no forces applied to the base and that just one point of each end-effectors contacts the environment. By manipulating the right hand side of the equation \eqref{eqn:dissipativeDyn} provides the \textit{dissipative equation of motion} of the robot
\begin{gather}
    \underbrace{\bM{K\T E M K}}_{\bM H(\bm \eta)}\ddot{\bm y} + \bM{K\T E}\bm c = \bM{\bar J}\T \bm f_{\textrm {ext}} + (\bM {\bar D} \bM {\bar G})\invT \bM {\bar E}\bm \tau_{\textrm{act}}. \label{dissipativeDynExpand}
    % \bM K_o = \bmat{\bm 1_6 &\bm 0 
    % \\ \bm 0 &(\bM{DG})^{-\mathrm{T}}}, \:\:\: \bm \tau_{\textrm{act}} = \bmat{\bm 0_{6\times1} \\ \bm{\tau_\phi}}
    % \bmat{\bm 0_{6\times 1} \\(\bM{DG})^{-\mathrm{T}}\bm\eta\bm\tau_\phi} 
    % \\
    % \bM{(H + H_{\textrm {rot}})}\ddot y = \bM J\T \bm f_{\textrm {ext}} + \bmat{\bm 0 \\(\bM{DG})^{-1}}u 
\end{gather}
where $\bM {\bar D} = \blkdiag(\bm 1_6, \bM D)$, $\bM {\bar G} = \blkdiag(\bm 1_6, \bM G)$, $\bM {\bar E} = \blkdiag(\bm 1_6, \bm \eta)$, and $\bm \tau_{\textrm{act}} = [\bm 0_{6\times1}\T ,\: \bm{\tau_\phi}\T]\T$ are block-diagonals and actuation torques for the reduced system. 
The result takes the form of conventional manipulator equations of motion \cite{ModernRobotics}, with the transmission efficiencies embedded into the inertia and Coriolis matrices, and generalized forces term. The inertia matrix of \eqref{dissipativeDynExpand} is non-symmetric. The appendix addresses this issue. 

\subsection{Application of the formulation}
In contrast with the conservative formulation, the dissipative equation of motion salients the concept of power flow in the system. There are different expressions if the actuators are driving the robot with torques $\bm \tau_\phi$ (forward-drive) or if the limbs are being accelerated by external forces $\bm f_{\textrm{ext}}$ (backward-drive). 

This concept of asymmetric transmission dynamics leads to separate design criteria if the robot actuators are expected to perform positive or negative work, and conventional design criteria related to the dynamics can be re-derived \cite{HandbookofRoboticsDesignCriteria, DynamicCapability}. We illustrate this characteristic by analyzing the design of a 2-DoF leg and compute its generalized inertia ellipsoid  \cite{AsadaInertiaEllipsoid}, its force capability or tip force bounds \cite{ModernRobotics}, and its ability to mitigate shock loads due to impacts between the foot and the ground. The latter is defined by the Impact Mitigation Factor (IMF) \cite{ImpactMitigation}, which ranges from zero to one and quantifies the inertial backdrivability of the mechanism. 

\subsubsection{Generalized Inertia Ellipsoid}
The task-space inertia or the generalized inertia ellipsoid (GIE) describes the inertia felt at the end-effector frame of a robot. Due to the dependency of the transmission dynamics on the direction of power flow, the end-effector inertia perceived by the actuators will be different than the inertia perceived by an external force backdriving the robot. The calculation of \textit{Backward-GIE} is identical to the conventional GIE, since the external force is not distorted by transmission efficiency. However, the propagation of the actuation torque is different in the case of \textit{Forward-GIE}. To accelerate the end-effector, the actuators exert a virtual task-space force $\hat{\bm f}_{\textrm{task}}$ given by 
\begin{gather}
    \bm{\tau_{\textrm{act}}} = {\bM{\bar G}}^{\mathrm{T}} {\bM {\bar D}}^{\mathrm{T}}  \bM{\bar J}\T{\bm {\hat{f}_{\textrm{task}}}}. \label{VirtualForce}
\end{gather}
Equations \eqref{VirtualForce} and \eqref{dissipativeDynExpand} provide the forward-GIE and requires the existence of the inverse of the Jacobian. If the Jacobian is non-square, the Moore-Penrose inverse, $(\:\: )^{+}$, is used instead. 
\begin{IEEEeqnarray*}{ccl}
    \textrm{FGIE} &=& \pmat{ \bM{\bar J} \bM H\inV (\bm \eta_f) \bM{\bar J}\T }\inV \!\! \bmat{{(\bM{\bar J} \bM{\bar D} \bM{\bar G})^{+}}\T \bM{\bar E(\bm \eta_f)} (\bM{\bar J} \bM{\bar D} \bM{\bar G} )\T}\inV \:\:\:\:\:,\\
    \textrm{BGIE} &=& \pmat{ \bM{\bar J} \bM H\inV (\bm \eta_b) \bM{\bar J}\T }\inV.
\end{IEEEeqnarray*}

Analogous to the forward and backward impedance describe in equations \eqref{eqn:ForwardImpedance} and \eqref{eqn:BackwardImpedance}, the FGIE and BGIE are larger in respect to the GIE due to the lower efficiency of transmission.

\subsubsection{Task-Space Force Capability}
The task-space force capability (FC) estimates the maximum contact force that the robot can produce at the end-effector. Conventionally, this concept shows that this force is limited by the maximum torque that the actuators can generate. However, \textit{we propose that it also depends on the mechanical efficiency of the robot's transmission.} To better understand this concept, assume that a legged robot semi-statically interacts with the ground to support it's body weight. In the forward driving case, the leg actuators must overcome the forces due to the robot mass plus the dissipative forces in the transmission. Hence, the lower the transmission's efficiency, the higher must be the actuator input effort to lift the robot. However, in the backward driving case, the gravitational torques due to the robot weight must drive the motors through the transmission. Interestingly, in this scenario, the lower efficiency (due to high friction) \textit{helps} the robot to passively support the its body. And thus, legged robots or industrial manipulators which employ highly geared and low efficiency actuators do not collapse under their own body weight when powered off. These systems behave like statues. This phenomena is captured by \textit{Asymmetric Force Capability} (AFC), which estimates the robot's ability to quasi-statically support or resist external forces. 
\begin{gather}
\textrm{AFC} = {\bM {(JDG)}^{+}}\T  \bm \eta \convhull({\bm \tau}_{{\phi}})
\end{gather}
Where we assume that the inverse of the Jacobian exists and tha the motor torque is bounded. The AFC only employs the limb's contact Jacobian $\bM J \in \R ^{3\times m}$, not that of the whole system.
\begin{figure}[t]
  \centering
  \includegraphics[width =1\linewidth]{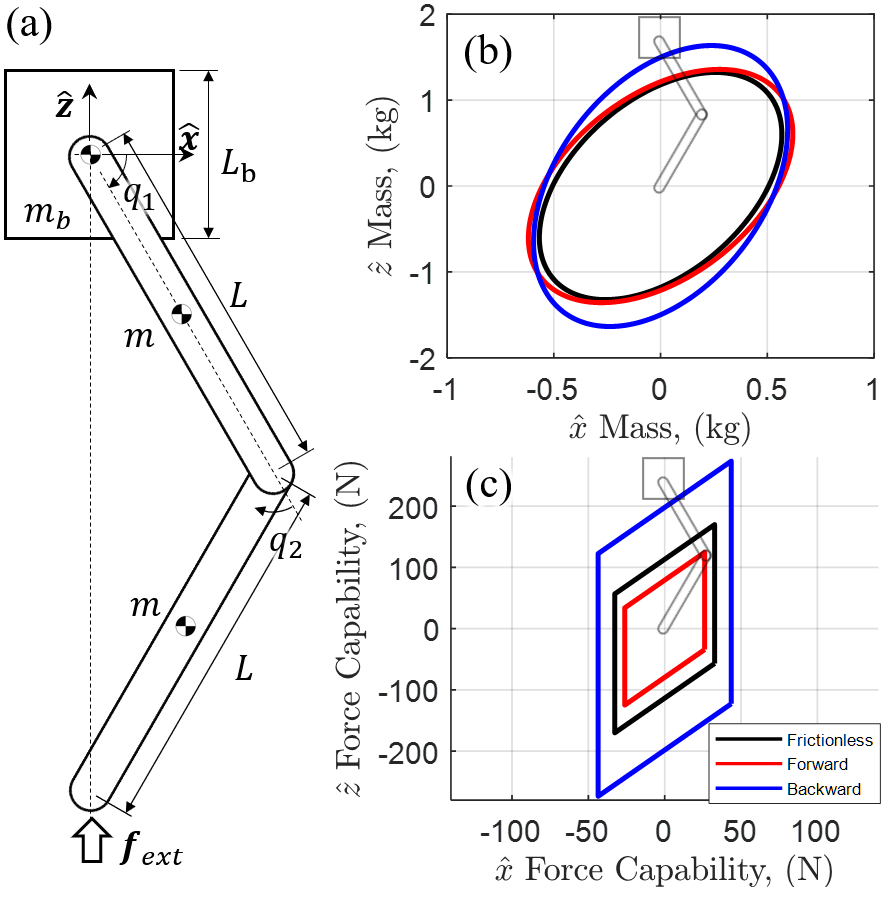}
  \caption{(a) The 2-DoF floating base robot for the design study (b) The forward and backward generalized inertia ellipsoids of the 2-DoF Leg (c) The normalized asymmetric force capabilities of the 2-DoF Leg}
  \label{Fig:GIEandFC}
\end{figure}
\begin{figure}[t]
  \centering
  \includegraphics[width =1\linewidth]{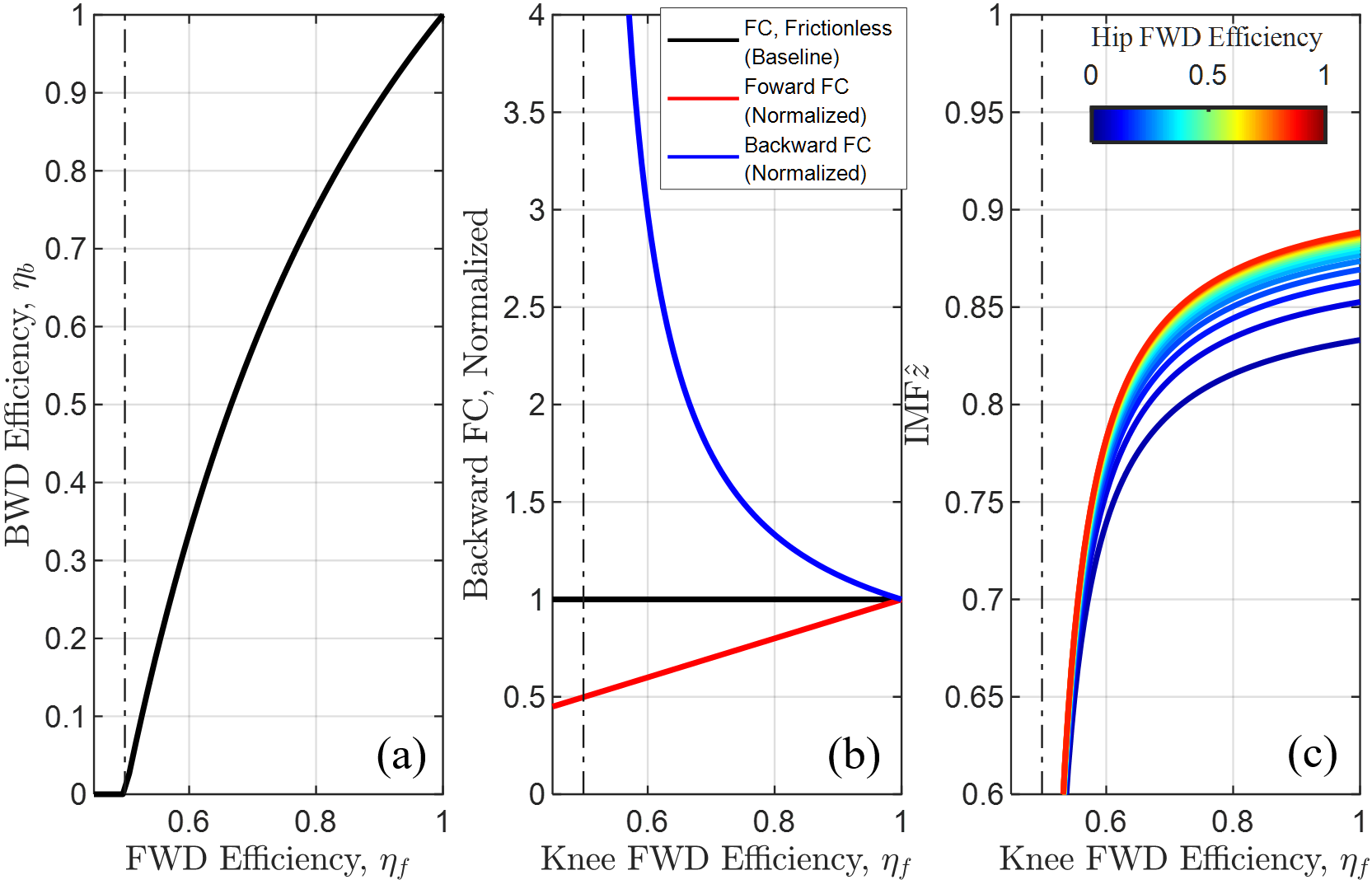}
  \caption{ (a) The relation between forward and backward efficiencies. The backward efficiency converges to zero at $\eta_f=0.499$. (b) The $\hat{z}$-directional forward and backward force capabilities normalized by force capability. (c) The $\hat{z}$-directional IMF of 2-DoF Leg}
  \label{Fig:IMFvsForceCapability}
\end{figure}

\subsection{Effect of dissipative dynamics}
We present two analysis of the design of a 2-DoF  planar leg composed of a serial actuation mechanism. First, we investigate the effect of transmission efficiencies on the task-space (foot) inertia matrix using the BGIE and FGIE, and the force capability using the AFC. Next, the AFC and the IMF in vertical $\hat z$ direction is computed to analyze the dynamic response of the leg to external contact forces. The AFC is normalized by conventional FC to measure the relative difference due to the internal losses in the transmissions. 

The leg, shown in Fig.\ref{Fig:IMFvsForceCapability}(a), is equipped with two identical motors that can exert up to $17\textrm{ N}\cdot \textrm{m}$ and are attached to the hip ($ q_1$) and the knee ($ q_2$) joints. The thigh and the sheen have the identical mass $m=0.4\textrm{ kg}$ and length $L=0.3\textrm{ m}$. Their center of mass are located at the links' midpoint. The base is modeled as a uniform planar square of dimensions $L_b= 0.4\textrm{ m}$ and mass $m_b=5\textrm{ kg}$. The base and the motors are at initially at rest, with joint angles, $q_1=q_2=60 \textrm{ deg}$. The speed reduction ratio is $1/20$ and the rotor inertia of the motor $6.4\times 10^{-5}\textrm{ kg}\cdot\textrm{m}^2$. We assume a typical forward efficiency of planetary gearboxes of $0.8$ and $0.7$ for the hip and knee transmission. However, the backward efficiency $\eta_b$ is a function of forward efficiency $\eta_f$ as displayed in Fig. {\ref{Fig:IMFvsForceCapability}}(b) and \eqref{eqn:eff}. The expression obtained from \cite{BilateralGear} is
\begin{equation}
    \eta_b = \begin{cases}
    \tfrac{2\eta_f-1+G^2}{(1-G^2)\eta_f +2G^2} & \Big ( \tfrac{1-G^2}{2}< \eta_f \le 1 \Big )
    \\
    \quad \quad \quad \quad 0 & \Big ( 0 < \eta_f \le \tfrac{1-G^2}{2} \Big )
    \end{cases}. \label{eqn:eff}
\end{equation}
We note that the backward efficiency is \textit{always} smaller than the forward efficiency, and converges to zero when $\eta_f \xrightarrow{} 0.499$. Herein, the index $j$ is dropped for both efficiencies and the reduction ratio, $G_j$.  

We observe the behavior of the asymmetric transmission dynamics in Fig. \ref{Fig:GIEandFC}. It is clear that both the FGIE and BGIE are always larger than GIE. Intuitively, the friction in the transmission makes it harder for the motors to move the robot and harder for external forces to backdrive the system. In contrast, the forward-FC (FFC) and backward-FC (BFC) show opposite tendencies. As the actuator efficiency decreases, the FFC linearly decreases while BFC diverges to infinity when $\eta_f\xrightarrow{}0.499$. Consequently, the leg can quasi-static withstand more substantial load. It is important to notice that both the task space inertia and the force capability are more significantly affected in the BWD case because the backward transmission efficiency is always inferior.

Finally, Fig. \ref{Fig:IMFvsForceCapability} shows the FC, FFC, BFC, and IMF of the foot in the vertical $\hat z$ direction. The result conveys the trade-off between the backward FC and the IMF. As the actuators' efficiency decrease, the BFC increases, while $\hat z-$IMF decreases. In other words, high mechanical losses in the transmissions allows the robot to sustain more substantial static forces, but also degrades the machine's ability to mitigate shock loads from impacts with the ground. 

\section{Conclusion}
This letter investigated how the dissipative forces in actuators and transmissions propagates to the dynamics of the whole robot. We present a framework which employs the mechanical efficiency, which is commonly provided by manufacturers, to augment the inertia, Coriolis, and generalized force terms in the equation of motion. 
We show how the individual efficiency of transmission influences the inertia felt at the end-effector and its capability of applying forces to and resisting disturbances from the environment. We expect that roboticists will use this formulation obtain an addition tunable variable to optimize the design of multi-body systems: the mechanical efficiency. 
For instance, designers may exploit the beneficial effects of low efficiency to gain more load-bearing capability, or to minimize the negative impacts of friction for dynamic tasks. 


%%%%%%%%%%%%%%%%%%%%%%%%%%%%%%%%%%%%%%%%%%%%%%%%%%%%%%%%%%%%%%%%%%%%%%%%%%%%%%%%

%%%%%%%%%%%%%%%%%%%%%%%%%%%%%%%%%%%%%%%%%%%%%%%%%%%%%%%%%%%%%%%%%%%%%%%%%%%%%%%%

%%%%%%%%%%%%%%%%%%%%%%%%%%%%%%%%%%%%%%%%%%%%%%%%%%%%%%%%%%%%%%%%%%%%%%%%%%%%%%%%
\section*{APPENDIX}

The inertia matrix ${\bM H(\bm \eta)=\bM {K} \T \bM{E} (\bm \eta) \bM{M K}}$ from equation \eqref{dissipativeDynExpand} is non-symmetric and thus, its positive-definiteness cannot be evaluated. This section presents a derivation to approximate a symmetric inertia matrix within small numerical error. The non-symmetric portion of the inertia matrix of equation (\ref{dissipativeDynExpand}) $\bM M_{\textrm{ns}}$ is converted to a symmetric form $\bM M_{\textrm {s}}$ with an algebraic manipulation, 
\begin{IEEEeqnarray*}{cCL}
    {\bM M}_{\textrm {ns}} = \bM{E M},\quad\quad\quad {\bM M}_{\textrm s} = \bM{\sqrt{E} M \sqrt{E}}.
\end{IEEEeqnarray*}

The validity of this conversion can be verified by calculating the error of kinetic energy. The difference in kinetic energy, $\Delta T= T_{\textrm s}-T_{\textrm {ns}}$, is calculated and divided by the actual kinetic energy, $T_{\textrm{ns}}$, to evaluate the error, $e=\tfrac{\Delta T}{T_{\textrm{ns}}}$,
\begin{IEEEeqnarray*}{rCl}
    \Delta T&=&T_{\textrm s}-T_{\textrm {ns}} =\mfrac{1}{2}\bm v\T \tilde{\bM M}_{\textrm s} \bm v- \mfrac{1}{2}\bm v\T \tilde{\bM M}_{\textrm {ns}} \bm v,
\end{IEEEeqnarray*}
where $\bm v$ is an arbitrary generalized velocity. It is assumed that all motors are either forward-driving or backward-driving. Hereafter, we consider forward-driving case. The first step of is to partition the inertia of the redundant system, $\bM M \in \R^{(6+2m)\times(6+2m)}$, into four sections, 
\begin{IEEEeqnarray*}{rCl}
    \bM M &=& \bmat{\bM M_1 & \bM M_c \\ \bM M^{\mathrm{T}}_c & \bM M_2},
\end{IEEEeqnarray*}
where $\bM M_1 \cin \R^{(6+m)\times(6+m)}$, $\bM M_2\cin \R^{m\times m} $ and $\bM M_c\in \R^{(6+m)\times(m)}$ accounts for the inertia with respect to the coordinate $\bm y=[\bm q_b \T, \bm q \T]\T$, $\bm \phi$, and the coordinate coupling of $\bm y$ and $\bm \phi$. Note that $\bM M$, and $\bM M_1$ are symmetric, and $\bM M_2$ is a diagonal matrix with the rotor inertia on the diagonal. The generalized velocity is dissected in accordance, $\bm v = [\bm v_1 \T, \bm v_2 \T]\T$. 

The kinetic energy difference $\Delta T$ is calculated as,
\begin{align*}
    & 2\Delta T = {\bm v}\T {\bM M}_{\textrm{s}} \bm v - {\bm v}\T {\bM M}_{\textrm{ns}} \bm v \\
    & = {\bm v}\T  \bigg(\bmat{\bM M_1 & \bM M_c \\ \bm \eta {\bM M}_c \T & \bm \eta \bM M_2} - \bmat{\bM M_1 & \bM M_c \sqrt{\bm \eta} \\ \sqrt{\bm \eta}\bM M_c \T & \sqrt{\bm \eta}\bM M_2 \sqrt{\bm \eta}}\bigg) \bm v
    \\
    & = \bm v\T \bmat{\bm 0 & \bM M_c (\bm 1 - \sqrt{\bm{\eta}}) \\ -(\bm 1 - \sqrt{\bm \eta})\sqrt{\bm \eta}\bM M^{\mathrm{T}}_c & \sqrt{\bm \eta}(\smashedunderbrace{\sqrt{\bm \eta} \bM M_2 - \bM M_2 \sqrt{\bm \eta}}{=\bm 0})}\bm v
    \\
    & = {\bm v}^{\mathrm{T}}_{2} \bmat{(\bm 1-\sqrt{\bm \eta})^2 \bM M^{\mathrm{T}}_c} \bm v_1 .  
\end{align*}

The upper bound of the error, $e$, is derived after a series of inequalities. We first define the coupled kinetic energy that accounts for the coupling of $\bm y$ and $\bm \phi$, $T^c_{\textrm{ns}}$, 

\begin{gather*}
   2 T_c ={\bm v}^{\mathrm{T}}_{1} \bM M_c \bm v_2 + {\bm v}^{\mathrm{T}}_{2} \bm \eta {\bM M}^{\mathrm{T}}_{c} \bm v_1 = {\bm v}^{\mathrm{T}}_{2} (\bm 1+\bm \eta){\bM M}^{\mathrm{T}}_{c} {\bm v}_{1} \\
   e_c:=\dfrac{\Delta T}{T_c}=\dfrac{{\bm v}^{\mathrm{T}}_{2} {(\bm 1- \sqrt{\bm \eta})^2 \bM M^{\mathrm{T}}_c} \bm v_1 }{{\bm v}^{\mathrm{T}}_{2} (\bm 1+\bm \eta){\bM M}^{\mathrm{T}}_{c} {\bm v}_{1}} =1-\underbrace{\dfrac{2\Sigma_{i=1}^m \sqrt{\eta_i} \alpha_i}{\Sigma_{i=1}^m {(1+\eta_i)\alpha_i}}}_{=:\gamma(\bm \eta)} .
\end{gather*}
% \vspace{-12pt}
where $\alpha_i\in\R$, $i\in\{1,...,m\}$. 
The coefficients of $\alpha_i$ of the denominator, $1+\eta_i$, are greater than those of the numerator, $\sqrt{\eta_i}$, and the coefficients are all positive. These properties bounds the absolute value of the $\gamma(\bm\eta)$,
\begin{align*}
       |\gamma(\bm \eta)| \ge \dfrac{2\Sigma_{i=1}^m \sqrt{\eta_i} |\alpha_i|}{\Sigma_{i=1}^m {(1+\eta_i)|\alpha_i|}} =:\delta(\bm \eta).  
\end{align*}
Now, the $\delta(\bm \eta)\in[0,1]$ is a strictly increasing function of $\eta_j\in[0,1]$, for all $j=\{1,...,m\}$. This can be proven by investigating all partial derivatives of the $\gamma$ with respect to all $\eta_j$,
\begin{align*}
    \pdif{\delta(\bm \eta)}{\eta_j} = \dfrac{|\alpha_j|\Sigma_{i=1}^m{(1+\eta_i -\sqrt{\eta_i})|\alpha_i|}}{(\Sigma_{i=1}^m {(1+\eta_i)|\alpha_i|})^2 \sqrt{\eta_j}} > 0 , 
\end{align*}
% where $\{k\}\cup\{l\}=\{1,\cdots,m\}$. 
Since the $\delta((\bm \eta))$ is increasing function of the $\eta_j$, it becomes clear that the $\delta$ has the smallest value when all efficiency values are replaced by the smallest efficiency, $\eta_{\textrm{min}}$,
\begin{align*}
    \delta(\bm \eta) \ge \delta(\eta_{\textrm{min}})= \dfrac{2\sqrt{\eta_{\textrm{min}}}}{1+\eta_{\textrm{min}}} . 
    % \\
    % \eta_{\textrm{min}} =  \min \big \{ \min_k \{\eta_{f}^k\}, \min_l\{\eta_b^l \} \big\} .
\end{align*}
Therefore, the error is bounded by a function of minimum efficiency. Notice that the result remains identical if $\eta_{\textrm{min}}$ is replaced by $\eta^{-1}_{\textrm{min}}$. In other words, the upper bound of the error has the same expression for both forward and backward-driving scenarios.
\begin{align*}
    \dfrac{\Delta T}{T}=e<e_c\le 1-\delta(\eta_{\textrm{min}}) = \dfrac{(1-\sqrt{\eta_{\textrm{min}}})^2}{1+\eta_{\textrm{min}}}.
\end{align*}
The case, $e_c \ge 2$, is assumed to be physically infeasible, and thus neglected.

Finally, it is proved that the upper bound of the error is a function of smallest efficiency in the whole system: 
\begin{gather*}
    e = \frac{\Delta T}{T_{\textrm {ns}}}  < \frac{\Delta T}{T^c_{\textrm {ns}}} <
    1- \dfrac{2 \sqrt{\eta_{\textrm{min}}}}{1+\eta_{\textrm{min}}}. 
\end{gather*} 
 For instance, the difference in kinetic energy is lesser than 3.2\% at 60\% efficiency and 0.6\% at 80\% efficiency. The error is relative to the coupled kinetic energy $T^c_{\textrm{ns}}$ which accounts for the coupled kinetic energy between the reduced velocity $\dot{\bm y}$ and rotors' angular velocity $\dot{\bm \phi}$.

%%%%%%%%%%%%%%%%%%%%%%%%%%%%%%%%%%%%%%%%%%%%%%%%%%%%%%%%%%%%%%%%%%%%%%%%%%%%%%%%

\bibliographystyle{plain}
\bibliography{bib_ywsim.bib}
% \begin{thebibliography}{99}
% \bibitem{FeatherstoneBook} Featherstone Book
% \bibitem{gearOptim1} gear Optim 1
% \bibitem{gearOptim2} gear Optim 2
% \bibitem{Yada} Yada
% \bibitem{BilateralEff} Bilateral Drive Gear—A Highly Backdrivable
% Reduction Gearbox for Robotic Actuators
% \bibitem{DirectionalEff} Wang
% \bibitem{WAMTownsend} Townsend, William T., and J. Kenneth Salisbury. "Mechanical design for whole-arm manipulation." Robots and Biological Systems: Towards a New Bionics?. Springer, Berlin, Heidelberg, 1993. 153-164.
% \bibitem{choiceMotorReducer}Giberti, H., S. Cinquemani, and G. Legnani. "Effects of transmission mechanical characteristics on the choice of a motor-reducer." Mechatronics 20.5 (2010): 604-610.
% \bibitem{safeSami} Haddadin, Sami, Alin Albu-Schäffer, and Gerd Hirzinger. "Requirements for safe robots: Measurements, analysis and new insights." The International Journal of Robotics Research 28.11-12 (2009): 1507-1527.
% \bibitem{impactMitigation} Wensing, Patrick M., et al. "Proprioceptive actuator design in the MIT Cheetah: Impact mitigation and high-bandwidth physical interaction for dynamic legged robots." Ieee transactions on robotics 33.3 (2017): 509-522.

% \end{thebibliography}
\end{document}